\documentclass[10pt,twocolumn]{article}

\usepackage[top=1in, bottom=1in, left=0.5in, right=0.5in]{geometry}

\usepackage{graphicx}
\usepackage{amsmath}
\usepackage{amssymb}
\usepackage{multirow}
\usepackage[pagebackref=true,breaklinks=true,letterpaper=true,colorlinks,bookmarks=false]{hyperref}
\usepackage{authblk}

\begin{document}

	\title
	{
		Object Detection with Pixel Intensity Comparisons Organized in Decision Trees
	}

	\author[$\dagger$]{Nenad Marku\v{s}}
	\author[$\dagger$]{Miroslav Frljak}
	\author[$\dagger$]{Igor S. Pand\v{z}i\'{c}}
	\author[$\ddagger$]{J\"orgen Ahlberg}
	\author[$\ddagger$]{Robert Forchheimer}

	\affil[$\dagger$]
	{
		\normalsize
		University of Zagreb,
		Faculty of Electrical Engineering and Computing,
		Unska 3, 10000 Zagreb, Croatia
	}
	\affil[$\ddagger$]
	{
		Link\"{o}ping University,
		Department of Electrical Engineering,
		SE-581 83 Link\"{o}ping, Sweden
	}

	\maketitle

	\begin{abstract}
		We describe a method for visual object detection based on an ensemble of optimized decision trees organized in a cascade of rejectors.
		The trees use pixel intensity comparisons in their internal nodes and this makes them able to process image regions very fast.
		Experimental analysis is provided through a face detection problem.
		The obtained results are encouraging and demonstrate that the method has practical value.
		Additionally, we analyse its sensitivity to noise and show how to perform fast rotation invariant object detection.
		Complete source code is provided at \url{https://github.com/nenadmarkus/pico}.
	\end{abstract}

	\section{Introduction}
		In computer vision, detection can be described as a task of finding the positions and scales of all objects in an image that belong to a given appearance class.
		For example, these objects could be cars, pedestrians or human faces.
		Automatic object detection has a broad range of applications.
		Some include biometrics, driver assistance, visual surveillance and smart human-machine interfaces.
		These applications create a strong motivation for the development of fast and accurate object detection methods.

		Viola and Jones \cite{viola-jones} have made object detection practically feasible in real world applications.
		This is due to the fact that systems based on their framework can process images much faster than previous approaches while achieving similar detection results.
		Still, some applications could benefit from faster detectors, and this was the main motivation for our research.
		We are interested in supporting a wide range of PC and mobile devices with limited processing power.
		Thus, to make our system practical in these applications, we are ready to sacrifice detection accuracy at the expense of better processing speeds and simplicity.

		In this paper, we describe an object detection framework which is able to process images very fast while retaining competitive accuracy.
		Basic ideas are described in Section \ref{sec:method}.
		Experimental analysis is provided in Section \ref{sec:facedetection}.
		Section \ref{sec:end} summarizes our findings and discusses future research directions.

		We have noticed (Feb. 2014) that an almost identical framework has been described by Liao et al. in a technical report \cite{almostidentical}.
		Here we acknowledge their work and state that our research has been performed completely independently.

	\section{Method}\label{sec:method}
		Our approach is a modification of the standard Viola-Jones object detection framework.
		The basic idea is to scan the image with a cascade of binary classifiers at all reasonable positions and scales.
		An image region is classified as an object of interest if it successfully passes all the members of the cascade.
		Each binary classifier consists of an ensemble of decision trees with pixel intensity comparisons as binary tests in their internal nodes.
		The learning process consists of a greedy regression tree construction procedure and a boosting algorithm.
		The details are given in the following subsections.

		\subsection{Decision tree for image based regression}\label{sec:dtree}
			To address the problem of image based regression, we use an optimized binary decision tree with pixel intensity comparisons as binary tests in its internal nodes.
			This approach was introduced by Amit and Geman in \cite{rand_trees_cv_1997}, and later successfully used by other researchers and engineers (for example, see \cite{kinect, ferns, puploc}).
			A pixel intensity comparison binary test on image $I$ is defined as
			\begin{equation}
				\text{bintest}(I; \mathbf{l}_1, \mathbf{l}_2) =
				\begin{cases}
					0,	&	I(\mathbf{l}_1)\leq I(\mathbf{l}_2)	\\
					1,	&	\text{otherwise}
					,
				\end{cases}
			\end{equation}
			where $I(\mathbf{l}_i)$ is the pixel intensity at location $\mathbf{l}_i$.
			Locations $\mathbf{l}_1$ and $\mathbf{l}_2$ are in normalized coordinates, i.e., both are from the set $[-1, +1]\times[-1, +1]$.
			This means that the binary tests can easily be resized if needed.
			Each terminal node of the tree contains a scalar that models the output.

			The construction of the tree is supervised.
			Training data is a set $\{(I_s, v_s, w_s): s=1, 2, \ldots, S \}$ where $v_s$ is the ground truth value associated with image $I_s$ and $w_s$ is its importance factor (weight).
			For example, in the case of binary classification, the ground truths represent class labels:
			Positive samples are annotated with $+1$ and negative samples with $-1$.
			The weights $w_s$ enable us to assign different importance value to each of these samples.
			This will prove important later.
			The binary test in each internal node of the tree is selected in a way to minimize the weighted mean squared error obtained when the incoming training data is split by the test.
			This is performed by minimizing the following quantity:
			\begin{equation}
				\text{WMSE} =
				\sum_{(I, v, w)\in C_0} w\cdot(v - \bar{v}_0)^2 + \sum_{(I, v, w)\in C_1} w\cdot(v - \bar{v}_1)^2
				,
				\label{eq:wmse}
			\end{equation}
			where $C_0$ and $C_1$ are clusters of training samples for which the results of binary test on an associated image were $0$ and $1$, respectively.
			Scalars $\bar{v}_0$ and $\bar{v}_1$ are weighted averages of ground truths in $C_0$ and $C_1$, respectively.
			As the set of all pixel intensity comparisons is prohibitively large, we generate only a small subset during optimization of each internal node by repeated sampling of two locations from a uniform distribution on a square $[-1, +1]\times[-1, +1]$.
			The test that achieves the smallest error according to equation \ref{eq:wmse} is selected.
			The training data is recursively clustered in this fashion until some termination condition is met.
			In our setup, we limit the depth of our trees to reduce training time, runtime processing speed and memory requirements.
			The output value associated with each terminal node is obtained as the weighted average of ground truths that arrived there during training.

			If we limit the depth of the tree to $D$ and we consider $B$ binary tests at each internal node, the training time is $O(D\cdot B\cdot S)$ for a training set containing $S$ samples, i.e., it is linear in all relevant parameters.
			This follows from the observation that each training sample is tested with $B$ pixel intensity comparisons for each internal node it visits on its path of length $D$ from the root node to its terminal node.
			A constructed tree needs $O(2^D)$ bytes for storage and its runtime speed scales as $O(D)$.

		\subsection{An ensemble of trees for binary classification}
			A single decision tree can usually reach only moderate accuracy.
			On the other hand, an ensemble of trees can achieve impressive results.
			We use the GentleBoost algorithm \cite{friedman-boosting}, a modification of the better known AdaBoost, to generate a discriminative ensemble by sequentially fitting a decision tree to an appropriate least squares problem.

			In order to generate an ensemble of $K$ trees from a training set $\{(I_s, c_s): s=1, 2, \ldots, S \}$, the algorithm proceeds in the following steps:
			\begin{enumerate}
				\item
					Initialize the weight $w_s$ for each image $I_s$ and its class label $c_s\in\{-1, +1\}$ as
					$$
						w_s=
						\begin{cases}
							1/P,	&	c_s=+1	\\
							1/N,	&	c_s=-1
						\end{cases}
					$$
					where $P$ is the total number of positive samples and $N$ is the total number of negative samples.

				\item
					For $k=1, 2, \ldots, K:$
					\begin{enumerate}
						\item
							Fit a decision tree $T_k$ by weighted least squares of $c_s$ to $I_s$ with weights $w_s$ (as explained in section \ref{sec:dtree}).
						\item
							Update weights:
							$$
								w_s=
								w_s \exp\left(-c_s T_k(I_s)\right)
								,
							$$
							where $T_k(I_s)$ denotes the real-valued output of tree $T_k$ on image $I_s$.
						\item
							Renormalize weights so they sum to $1$.
					\end{enumerate}

				\item
					Output ensemble $\{T_k:\; k=1, 2, \ldots, K\}$.
			\end{enumerate}

			During runtime, the outputs of all trees in the ensemble are summed together and the obtained value is thresholded in order to obtain a class label.
			We can achieve different ratios of true positives to false positives by varying the threshold.
			This proves important in building an efficient detector, as described in the next section.

		\subsection{Detection as image scanning with a cascade of binary classifiers}
			Without any \textit{a priori} knowledge, we have to systematically scan the image with our binary classifier at all different positions and scales in order to find the objects of interest.
			As this is computationally demanding, we follow the proposal of Viola and Jones.
			The basic idea is to use multiple classification stages with increasing complexity.
			Each stage detects almost all objects of interest while rejecting a certain fraction of non-objects.
			Thus, the majority of non-objects are rejected by early stages, i.e., with little processing time spent.

			In our case, each stage consists of an ensemble of trees.
			Early stages have fewer trees than the later ones.
			The detection rate of each stage is regulated by adjusting the output threshold of its ensemble.
			Each stage uses the soft output ("confidence") of the previous stage as additional information to improve its discriminability.
			This is achieved by progressively accumulating the outputs of all classification stages in the cascade (similar to \cite{boosting-chain}).

		\subsection{Clustering raw detections}
			As the final detector is robust to small perturbations in position and scale, we expect multiple detections around each object of interest.
			These overlapping detections are clustered together in a post-processing step as follows.

			We construct an undirected graph in which each vertex corresponds to a raw detection.
			Two vertices are connected if the overlap of their corresponding detections is greater than $30\%$:
			\begin{equation}
				\frac{D_1\cap D_2}{D_1\cup D_2}>
				0.3
				.
			\end{equation}
			Next, we use the depth-first search to find the connected components in this graph.
			Raw detections within each component are combined in a single detection by averaging the position and scale parameters.
			This simple clustering procedure does not require the number of clusters to be set in advance.

	\section{Face detection experiments}\label{sec:facedetection}
		It is not obvious that the described framework can give pleasing results in practice.
		Experimental analysis is provided through the face detection problem.
		A thorough survey of the field can be found in \cite{majki-survey}.

		\subsection{Training}
			For the purpose of training, we use the AFLW \cite{aflw} dataset and the one provided by Visage Technologies (\url{http://www.visagetechnologies.com/}).
			Both consist of face images annotated with approximate locations of the eyes.
			These are used to estimate the position and scale of each face.
			We extract around $20\,000$ frontal faces from these datasets and generate $15$ positive training samples from each frontal face by small random perturbations on scale and position.
			We have observed in our preliminary experiments that this makes the detector more robust to aliasing and noise.
			Overall, this results in around $300\,000$ positive samples.
			For learning of each stage we extract $300\,000$ negatives from a large set of images that do not contain any faces by collecting the regions that were not discarded by any of the previously learned stages.

\begin{figure}
				\center
				\includegraphics[scale=0.45]{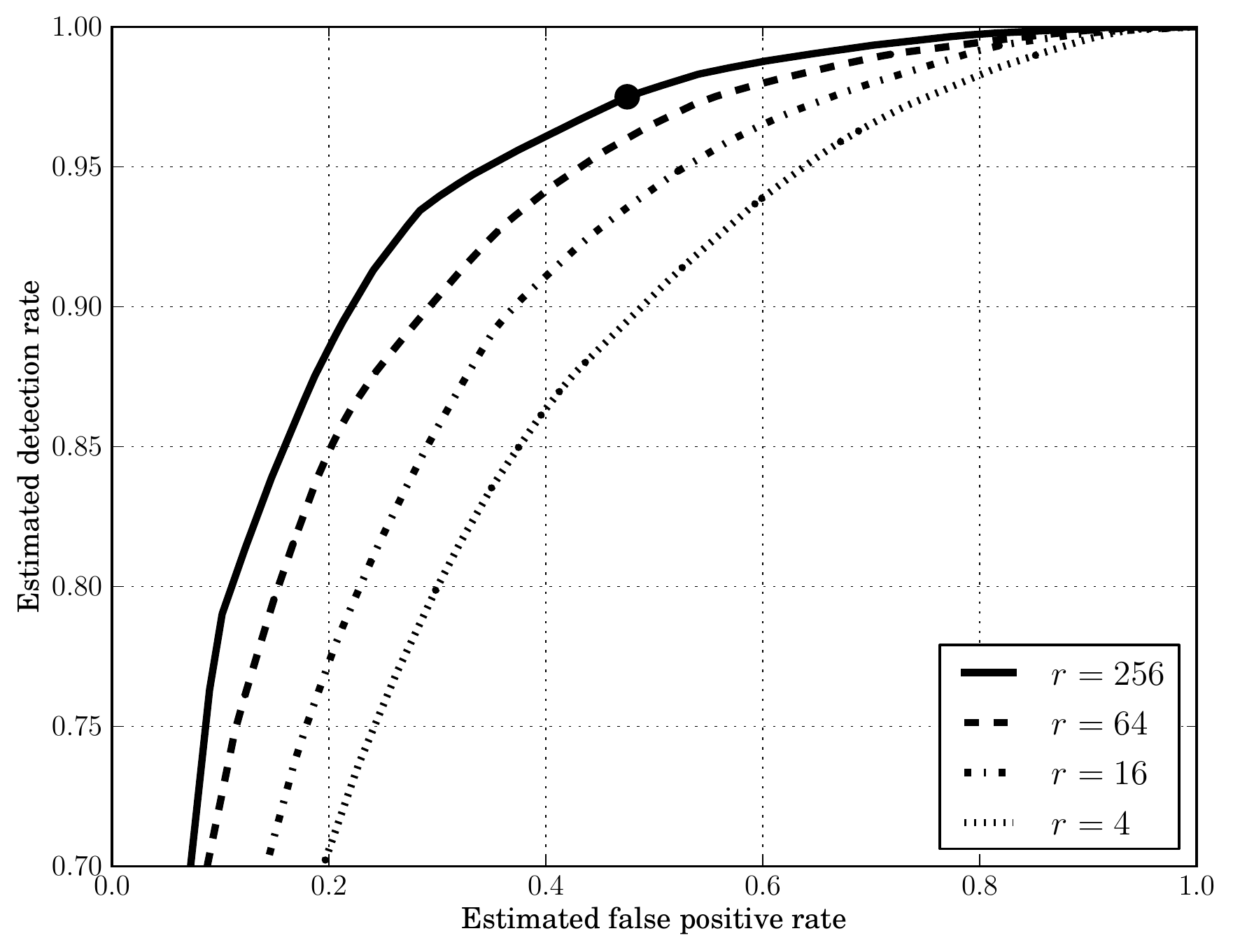}

				\caption
				{
					The ROC curves of the first stage of the cascade for different number of binary tests considered in the optimization of each internal tree node, $r$.
					Circular marker represents the point on which the stage operates at runtime.
				}
				\label{fig:first-stage-roc}
			\end{figure}

			The parameters of the learning process have to be set in advance.
			In our experiments, we tune them to produce a detector which is able to process images rapidly, as this is our main goal.
			We fix the depth of each tree to $6$ and use $20$ classification stages.
			Each classification stage has a predefined number of trees and detection rate.
			We consider $256$ binary tests during the optimization of each internal tree node.
			This optimization process considerably improves the performance of the cascade.
			For example, the first stage consists of a single tree.
			Its parameterized receiver operator characteristic (ROC) curve can be seen in Figure \ref{fig:first-stage-roc}.
			This experiment also implies that using randomized ferns \cite{ferns} in this framework leads to inferior processing speeds at runtime since ferns discard less negatives for the same stage complexity and detection rate.
			Similar conclusions can be made for other stages of the cascade.
			Some parameters and results of the learning process are reported in Table \ref{tbl:stage-params}.
			\begin{table*}
				\begin{center}
				{
				\begin{tabular}{| l || c | c | c | c | c | c | c | c | c | c | c |}
					\hline
					num. trees	&	$1$	&	$2$	&	$3$	&	$4$	&	$5$	&	$10$	&	$20$	&	$20$	&	\ldots	&	$20$	&	$20$	\\
					\hline
					TPR [\%]	&	$97.5$	&	$98.0$	&	$98.5$	&	$99.0$	&	$99.5$	&	$99.7$	&	$99.9$	&	$99.9$	&	\ldots	&	$99.9$	&	$99.9$	\\
					\hline
					FPR [\%]	&	$46.4$	&	$32.3$	&	$20.5$	&	$35.4$	&	$44.7$	&	$36.8$	&	$29.5$	&	$31.6$	&	\ldots	&	$55.2$	&	$57.5$	\\
					\hline
				\end{tabular}
				}
				\end{center}
				\caption
				{
					Number of trees, true positive rate (TPR) and estimated false positive rate (FPR) for some stages.
				}
				\label{tbl:stage-params}
			\end{table*}
			The overall detection rate on the training set is approximately $0.92$ for the estimated false positive rate of $10^{-7}$.
			Note that this apparently low detection rate does not mean poor performance in practice since we generated $15$ randomized samples for each frontal face image in the available datasets.
			Also, we use the scanning window approach during runtime and expect multiple detections for each encountered face.

			The learning takes around 30 hours on a modern PC with $4$ cores and $16$GB of RAM.
			Most of the computation effort goes to the sampling of negatives for learning of each new classification stage.
			The learned cascade uses less than $200$ kB of storage.

		\subsection{Analysis of accuracy and processing speed}
			To put our system into perspective, we compare its detection rate, false positive rate and processing speed to the tried-and-true face detection implementations available in OpenCV (\url{http://opencv.org/}, version 2.4.3).
			The first one is based on Haar-like features (standard Viola-Jones framework, see \cite{analysis-lienhart}) and the other on local binary patterns (LBPs, see \cite{mb-lbps}).
			We would like to note that there are certain limitations to the experiments that follow and we will not be able to provide absolute measures how the methods compare.
			The idea is to compare the implementation of our method to the baseline provided by OpenCV.

\begin{figure}
				\center
				\includegraphics[scale=0.45]{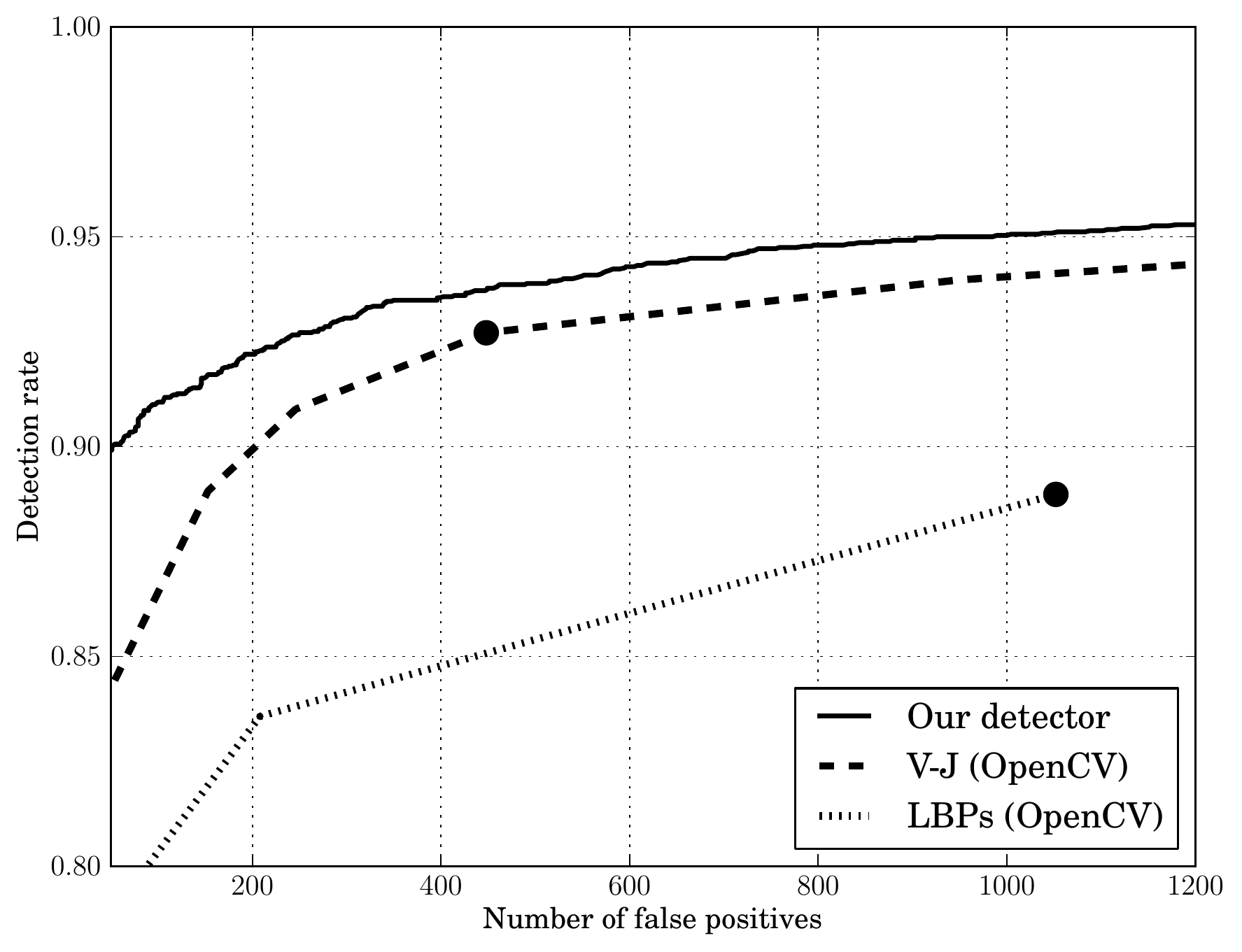}

				\caption
				{
					ROC curves for the GENKI-SZSL/NO-FACES-1 experiment.
				}
				\label{fig:genki-roc}
			\end{figure}

\begin{figure}
				\center
				\includegraphics[scale=0.45]{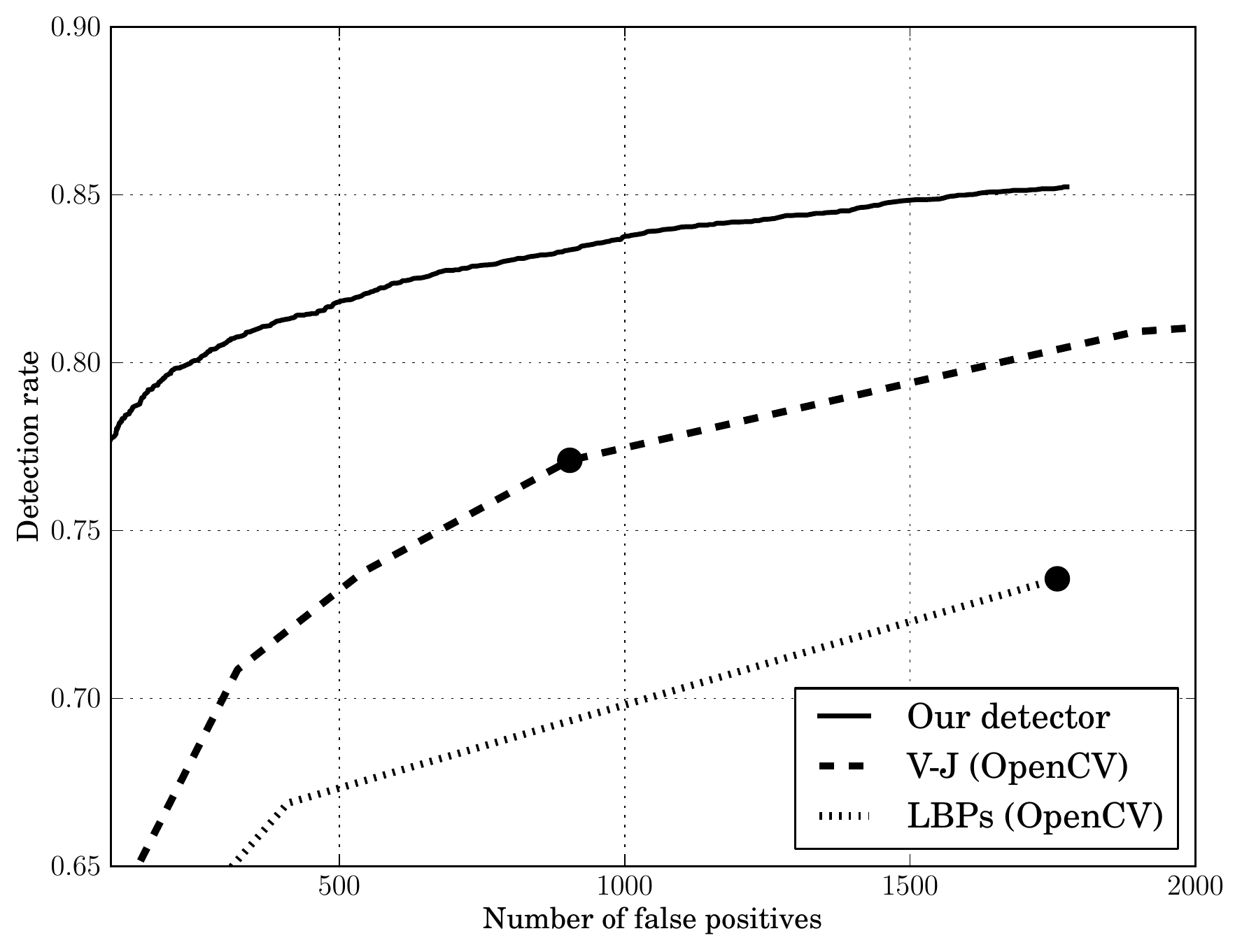}

				\caption
				{
					ROC curves for the CALTECH-FACES/NO-FACES-2 experiment.
				}
				\label{fig:caltech-roc}
			\end{figure}

			To evaluate the detection rates, we use the GENKI-SZSL \cite{genki-szsl} and CALTECH-FACES \cite{caltech-faces} datasets.
			The datasets contain $3500$ and $10\,000$ annotated faces, respectively.
			We report the number of false positives on two large sets of images that do not contain any faces, NO-FACES-1 and NO-FACES-2.
			All detectors start the scan with a $24\times 24$ pixel window and enlarge it by $20\%$ until it exceeds the size of the image.
			For a given scale, our system scans the image by shifting the window by $10\%$ of its size.
			The ROC curves can be seen in Figures \ref{fig:genki-roc} and \ref{fig:caltech-roc}
			(circular markers represent the recommended operating points for OpenCV detectors in real-time applications, as found in the documentation).
			We conclude that our system slightly outperforms the other two detectors in terms of accuracy.
			Of course, there is always the problem of dataset bias \cite{dataset_bias} and unknown implementation details, i.e., we cannot conclude which method is more accurate in general situations based just on these limited experiments.

			The V-J detector scans the NO-FACES-1 set in 602 seconds and the LBP-based detector in 240 seconds.
			Our system needs 111 seconds for the task.
			The reported times are on a 2.5GHz machine. 
			We are interested in a more realistic setup:
			Scan a $640\times 480$ image starting with a $100\times 100$ window that is enlarged by $20\%$ until it exceeds the size of the image.
			This situation is commonly encountered in real-world applications such as video conferencing or face tracker initialization.
			The processing speeds are reported in Table \ref{tbl:speed}.
			\begin{table*}
				\center
				\begin{tabular}{|c||c||c|c|c|}
					\hline
					\multirow{2}{*}{Device} & \multirow{2}{*}{CPU} & \multicolumn{3}{|c|}{Time [ms]} \\
					\cline{3-5}
					&	&	Our detector	&	V-J (OpenCV)	&	LBPs (OpenCV)	\\
					\hline
					\hline
					PC1	&	3.4GHz Core i7-2600	&	$2.4$	&	$16.9$	&	$4.9$	\\
					\hline
					PC2	&	2.53GHz Core 2 Duo P8700	&	$2.8$	&	$25.4$	&	$6.3$	\\
					\hline
					iPhone 5	&	1.3GHz Apple A6	&	$6.3$	&	$175.3$	&	$47.3$	\\
					\hline
					iPad 2	&	1GHz ARM Cortex-A9	&	$12.1$	&	$347.6$	&	$103.5$	\\
					\hline
					iPhone 4S	&	800MHz ARM Cortex-A9	&	$14.7$	&	$430.3$	&	$129.2$	\\
					\hline
				\end{tabular}
				\caption
				{
					Average times required to find faces larger than $100\times 100$ pixels in a $640\times 480$ image.
				}
				\label{tbl:speed}
			\end{table*}
			Bear in mind that the implementation available in OpenCV is highly optimized for PCs, i.e., it uses SIMD instructions, multi-core processing and cache-optimized data structures.
			Its poor performance on mobile devices can be explained by limited hardware support for floating point operations on ARM architectures
			\footnote{We are not the first to notice this problem with OpenCV. For example, see \url{http://www.computer-vision-software.com/blog/2009/04/fixing-opencv/} (accessed on 29th of October, 2013).}.
			Our implementation is written in pure \texttt{C}, without much time spent on tweaking for processing speed.
			Also, all processing is done in a single thread, i.e., uses a single CPU core.

			We conjecture that it is possible to obtain even better results with more advanced cascade construction/optimization techniques (for example, see\cite{soft, waldboost}).
			This kind of experiments were beyond the scope of the present paper and we plan to perform some of them in the future.
			Also, note that we can increase the accuracy of our face detector at the cost of processing speed by scanning the images more densely, for example, by enlarging the scanning window by $10\%$ instead of $20\%$ when changing the detection scale.
			The choice of these parameters depends on the targeted application.

		\subsection{Comparison with other methods}
\begin{figure}
				\center
				\includegraphics[scale=0.45]{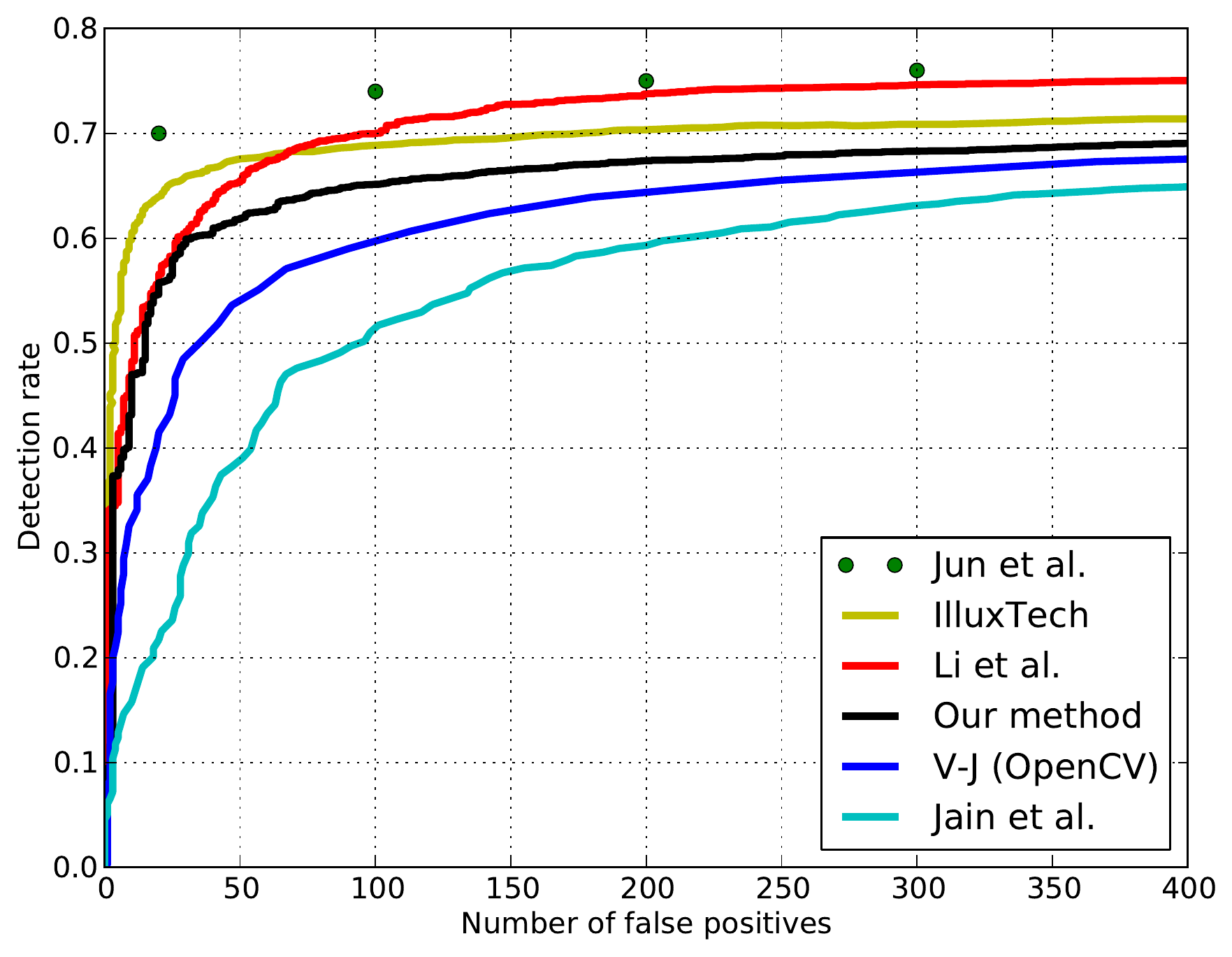}

				\caption
				{
					Discrete ROC curves for different face detection systems on the FDDB dataset.
				}
				\label{fig:fddb-discroc}
			\end{figure}

\begin{figure}
				\center
				\includegraphics[scale=0.45]{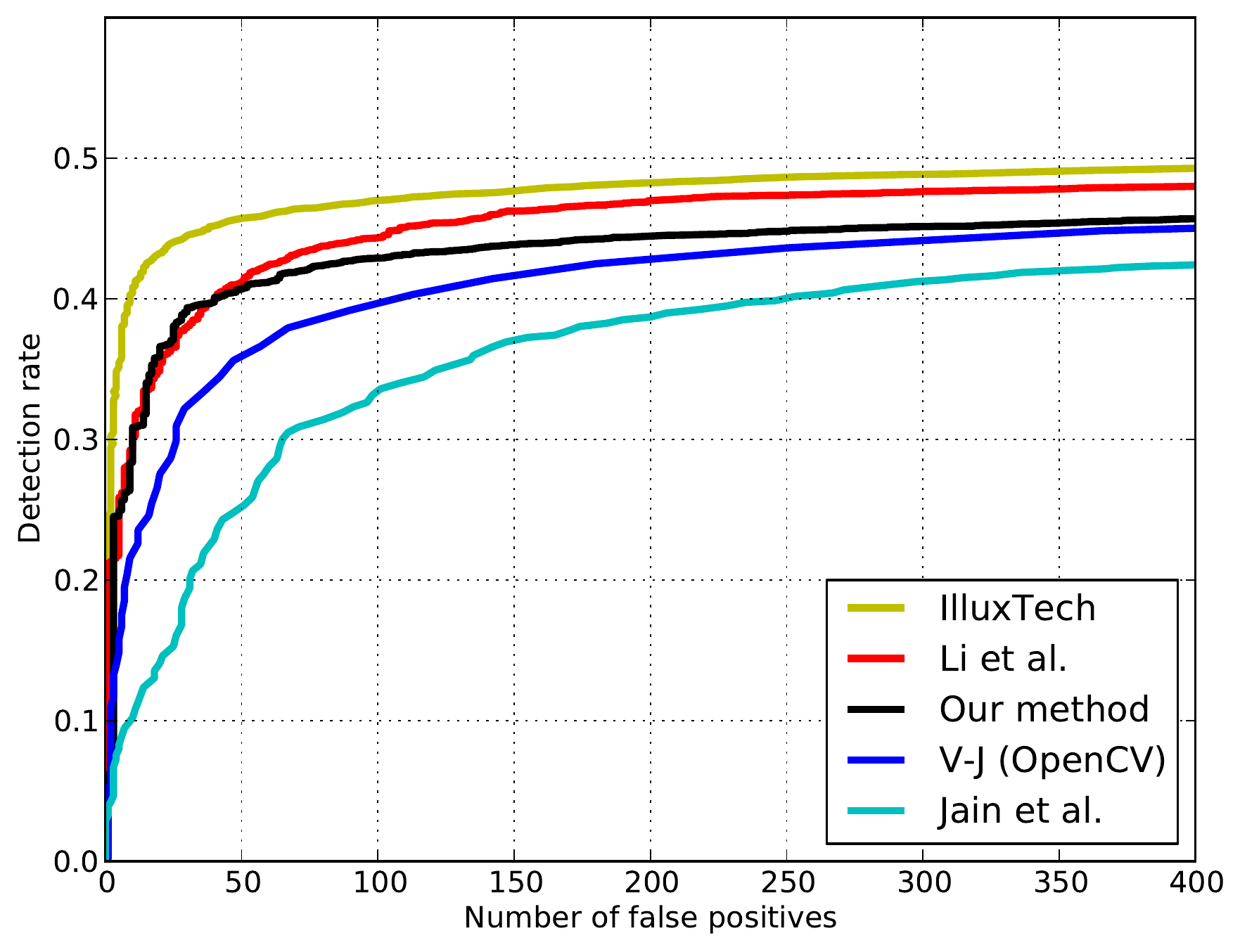}

				\caption
				{
					Continuous ROC curves for different face detection systems on the FDDB dataset.
				}
				\label{fig:fddb-controc}
			\end{figure}

			A common face detection accuracy benchmark is the FDDB dataset \cite{fddb}.
			It contains $5171$ faces acquired under unconstrained conditions.
			Some annotated faces are in out-of-plane rotated positions and frontal face detectors will not be able to find them.
			As we do not have access to the source code/binaries of the face detection systems considered to be state-of-the-art, we rely on the results presented in the papers describing them.
			Some are summarized at \url{http://vis-www.cs.umass.edu/fddb/results.html} and the other were obtained by visual inspection of the available accuracy plots.
			We use the protocol from \cite{fddb} to compare our system to the methods by Jun et al. \cite{lgp}, Li et al. \cite{surf}, Jain et al. \cite{vjgpr} and a commercial system provided by IlluxTech (\url{http://illuxtech.com/}).
			The discrete and continuous ROC curves can be seen in figures \ref{fig:fddb-discroc} and \ref{fig:fddb-controc}.
			We can see that \cite{lgp, surf} and IlluxTech outperform our system on this dataset in terms of accuracy.
			Li et al. \cite{surf} and Jun et al. \cite{lgp} compare the processing speeds of their systems to the V-J frontal face detector from OpenCV.
			All measurements were performed on modern personal computers.
			It is difficult to conjecture from the available data how the mentioned systems compare to ours because we do not know which parameters their authors used for the V-J detector in their experiments.
			If we assume that they used the same parameters as we did, this makes our system faster than Li et al. \cite{surf} by approximately $4.5$ times, and at the same speed as Jun et al. \cite{lgp}.

			Some of object detection research focuses on building ever more accurate systems, even at the expense of processing speed (for example, deformable part-based models with gradient orientation histogram features, see \cite{latentsvm, dpms}).
			As far as we know, the system presented by Yan et al. \cite{state-of-the-art-accuracy} obtains state-of-the-art face detection accuracy results on FDDB.
			The system uses multiple detectors, each trained to detect faces in a predefined range of out-of-plane rotations.
			Thus, we did not include these results in figures \ref{fig:fddb-discroc} and \ref{fig:fddb-controc}.
			The authors report a processing speed of around $20$ frames per second for frontal face detection in $640\times 480$ images on a modern personal computer.
			This suggests that their system would not achieve acceptable performance on mobile devices and other hardware with limited processing power.
			More in depth comparisons are not possible since we do not have access to the implementation of their method.
			Also, we conjecture that neural network-based object detection systems will obtain even better results in the future as neural networks started to outperform other machine learning methods on common benchmarks \cite{hinton}.
			However, neural networks are usually slow at runtime as they require a lot of floating point computations to produce their output, which is particularly problematic on mobile devices.
			To conclude, our opinion is that similar systems are not suitable for the same kind of applications as the face detector described in this paper.

		\subsection{Sensitivity to noise}
\begin{figure}
				\center
				\includegraphics[scale=0.45]{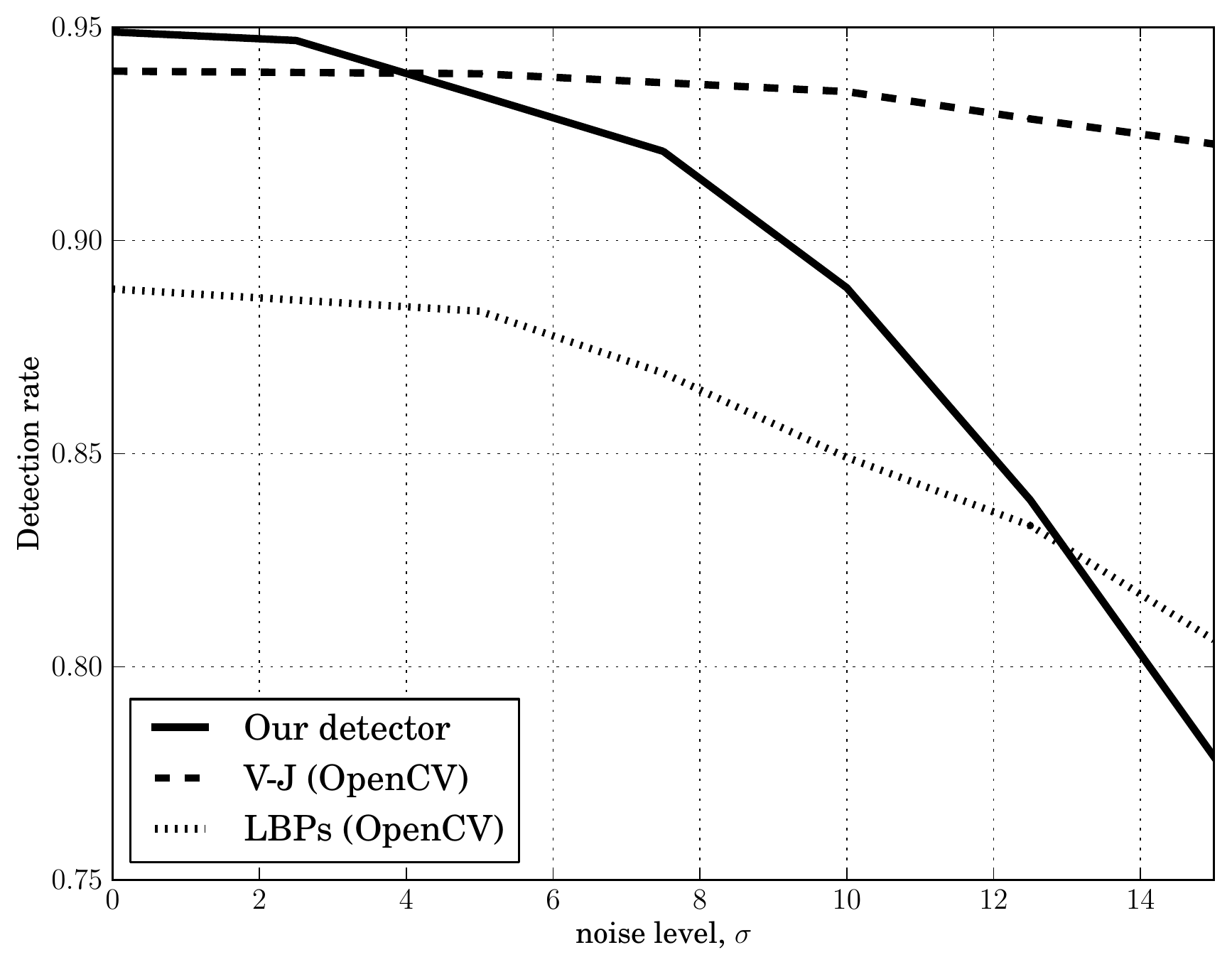}

				\caption
				{
					Detection rate on the GENKI-SZSL dataset for different noise levels.
				}
				\label{fig:genki-noise}
			\end{figure}

			It is reasonable to assume that our system performs poor in the presence of high noise levels due to simplicity of the used binary tests.
			On the other hand, the features used by OpenCV detectors should be robust in these circumstances as they are based on region averaging, which corresponds to low-pass filtering.
			We use the additive (uncorrelated) Gaussian noise model to quantitatively evaluate these effects:
			A sample from a Gaussian distribution with zero mean and standard deviation $\sigma$ is added to the intensity of each image pixel.
			An experiment on the GENKI-SZSL dataset is reported in Figure \ref{fig:genki-noise}.
			We see that the detection rate of our system degrades significantly as the noise intensity increases.
			These adverse effects can be reduced by applying a low-pass filter prior to detection.
			We have not found this to be necessary as the noise levels in this experiment are uncommon for modern cameras, even on mobile devices.
			The images in the GENKI-SZSL dataset are already noisy and contain compression artefacts, i.e., they are representative of the ones encountered in real-world face detection applications.

			Note that the presented experiment indicates that other systems based on similar features could be sensitive to high noise levels as well.
			An example of such a commercial system is the Microsoft Kinect human pose recognizer, described in \cite{kinect}.

		\subsection{Detecting rotated faces}
			In some applications we are interested in object detection under general planar rotations.
			A simple solution is to scan the image at multiple orientations.
			In our case, this can be performed without image resampling as pixel intensity comparison binary tests can be easily rotated for a given angle
			(in our implementation, we use precomputed look-up tables as this proved faster than evaluating trigonometric functions at runtime).
			It is not immediately clear if this leads to acceptable performance since it could result in poor processing speeds and/or large number of false positives.
			We use the previously learned face detector and provide experimental analysis.

\begin{figure}
				\center
				\includegraphics[scale=0.45]{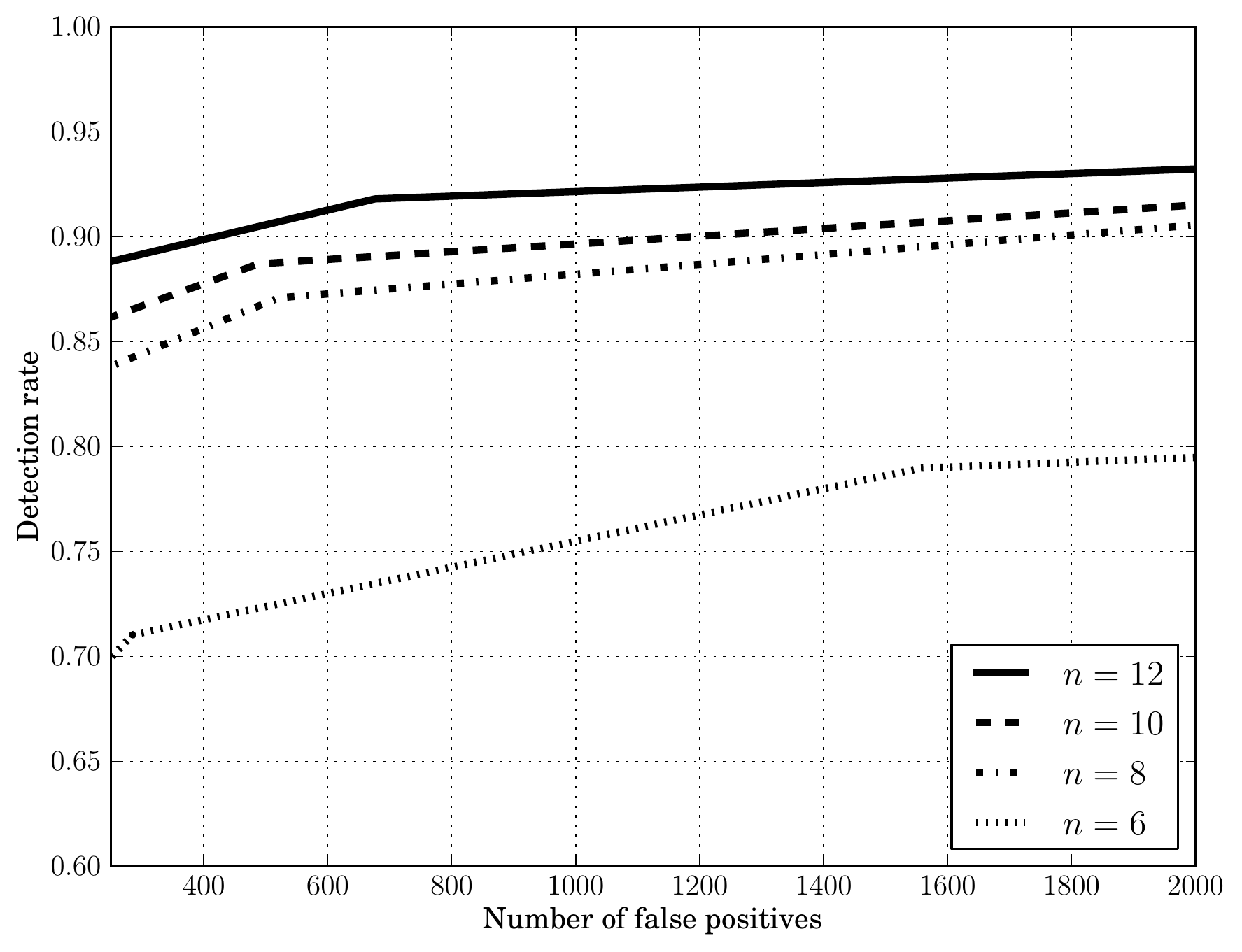}

				\caption
				{
					ROC curves for different number of considered face orientations, $n$, during the scan of each image.
				}
				\label{fig:rot-roc}
			\end{figure}

			To investigate the detection accuracy, we rotate each image found in the GENKI-SZSL dataset for a random angle sampled uniformly from the $[0, 2\pi)$ interval.
			We report false positives on the NO-FACES-1 set.
			Results can be seen in Figure \ref{fig:rot-roc}.
			By comparing the ROC curves to the ones in Figure \ref{fig:genki-roc} we can see that the accuracy of the approach is comparable to the OpenCV's LBP-based frontal face detector.
			Processing speeds can be seen in Table \ref{tbl:rotspeed}.
			\begin{table}
				\center
				\begin{tabular}{|c||c|c|c|c|}
					\hline
					\multirow{2}{*}{Device} & \multicolumn{4}{|c|}{Time [ms]} \\
					\cline{2-5}
					&	$n=6$	&	$n=8$	&	$n=10$	&	$n=12$	\\
					\hline
					\hline
					PC1	&	15.8	&	20.6	&	25.8	&	31.25	\\
					\hline
					PC2	&	21.6	&	26.8	&	34.5	&	42.9	\\
					\hline
				\end{tabular}
				\caption
				{
					Average times required to process a $640\times 480$ pixel image at $n$ orientations, attempting to find faces larger than $100\times 100$ pixels at all possible planar rotations.
				}
				\label{tbl:rotspeed}
			\end{table}
			These results demonstrate that our system can perform rotation invariant face detection with reasonable accuracy in real-time using a single core of a modern personal computer.

		\subsection{Qualitative results}
\begin{figure*}
				\center

				\scalebox{1.2}
				{
					\includegraphics[height=5.15cm]{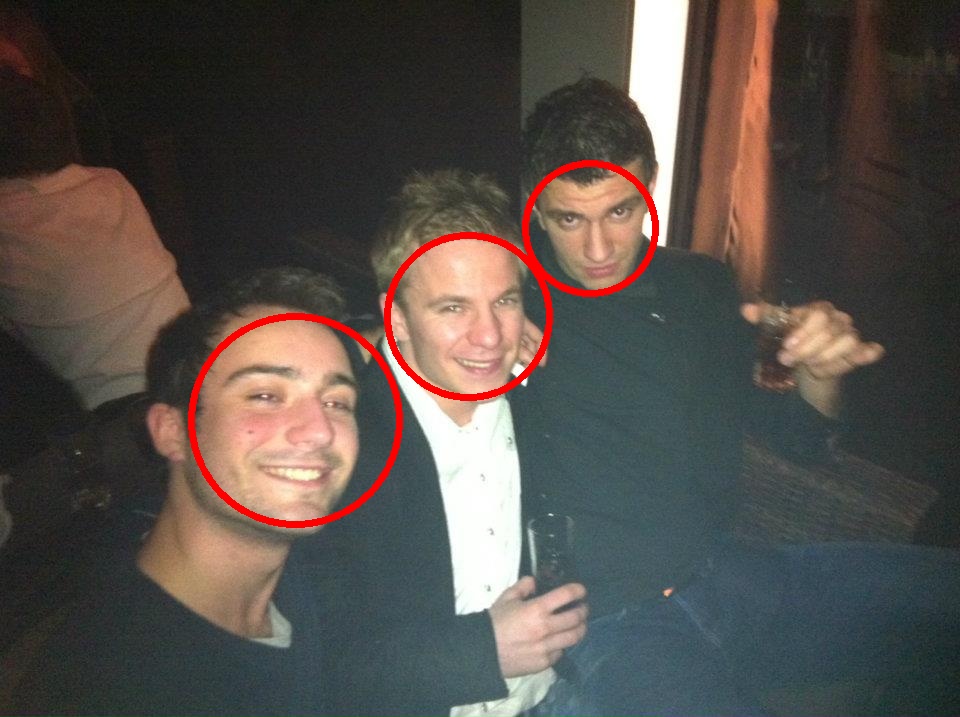}
					\includegraphics[height=5.15cm]{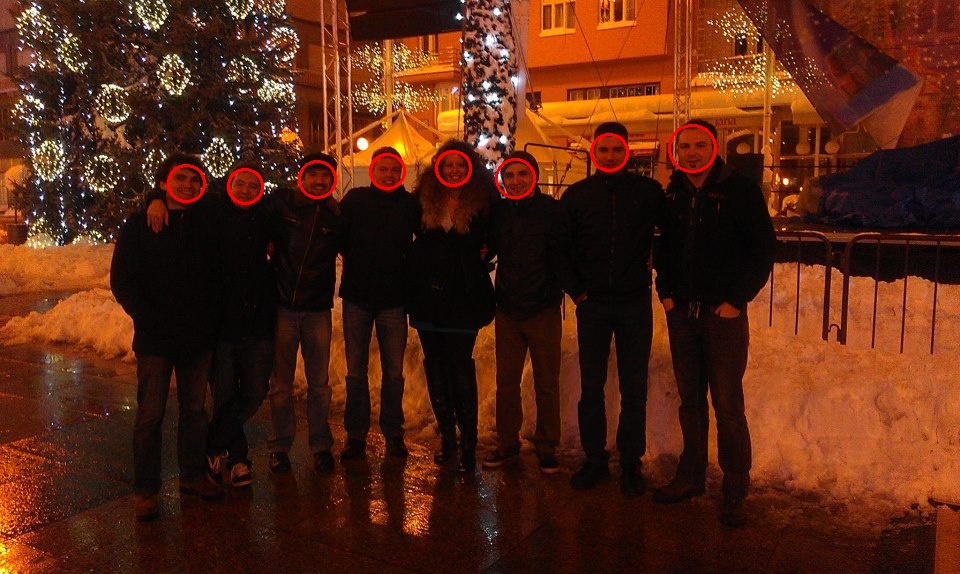}
				}

				\scalebox{1.2}
				{
					\includegraphics[height=2.9cm]{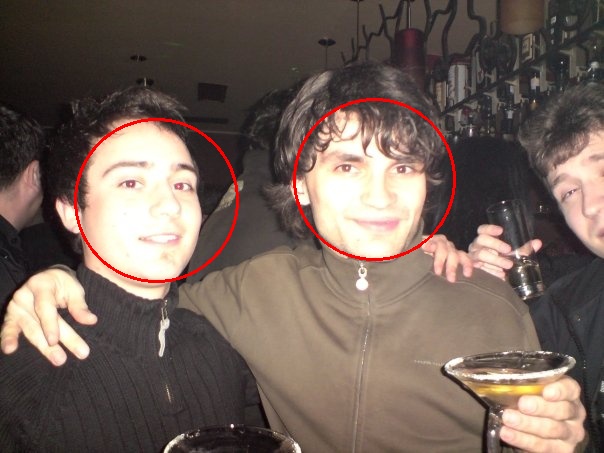}
					\includegraphics[height=2.9cm]{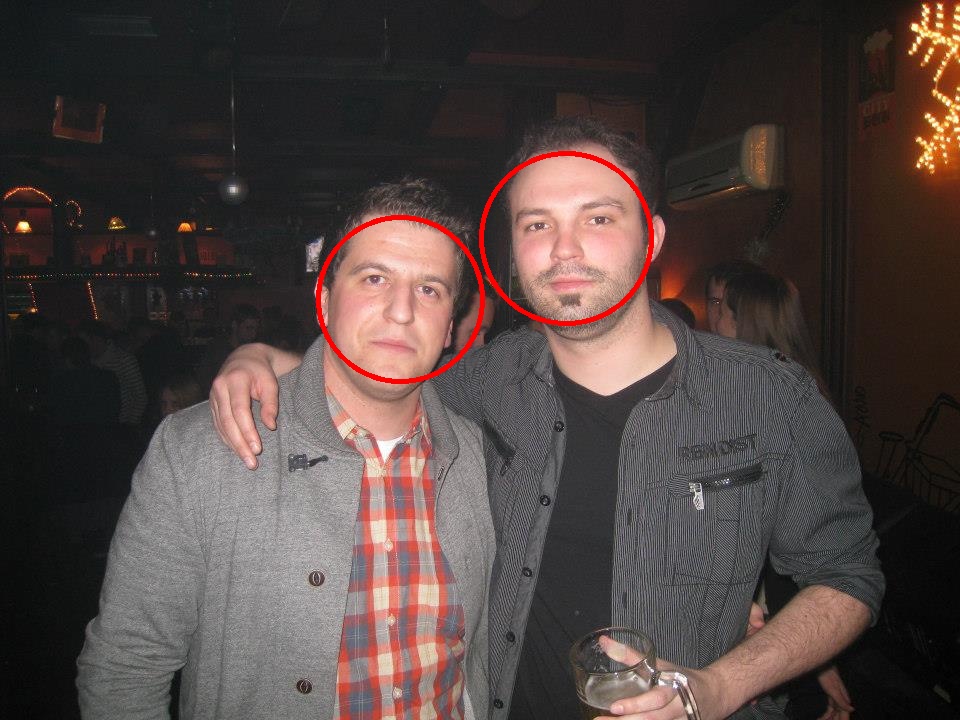}
					\includegraphics[height=2.9cm]{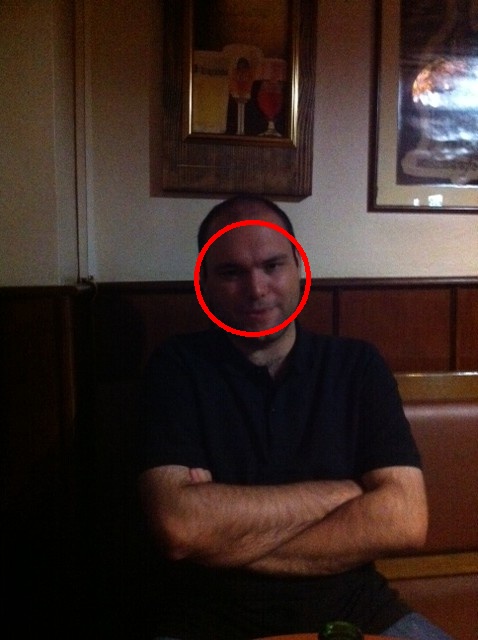}
					\includegraphics[height=2.9cm]{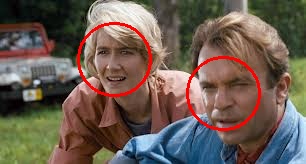}
				}

				\scalebox{1.2}
				{
					\includegraphics[height=2.7cm]{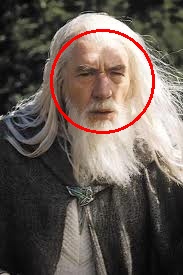}
					\includegraphics[height=2.7cm]{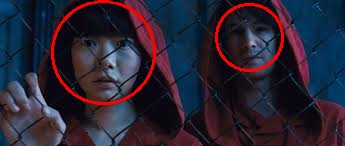}
					\includegraphics[height=2.7cm]{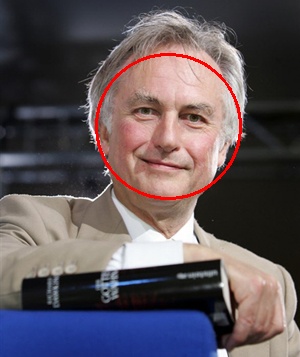}
					\includegraphics[height=2.7cm]{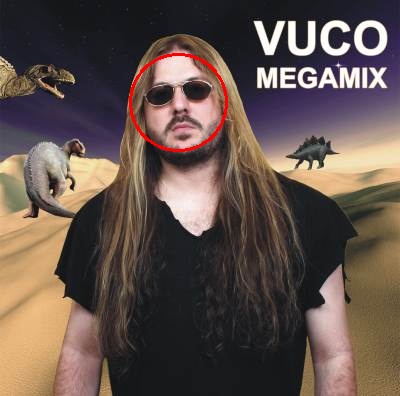}
					\includegraphics[height=2.7cm]{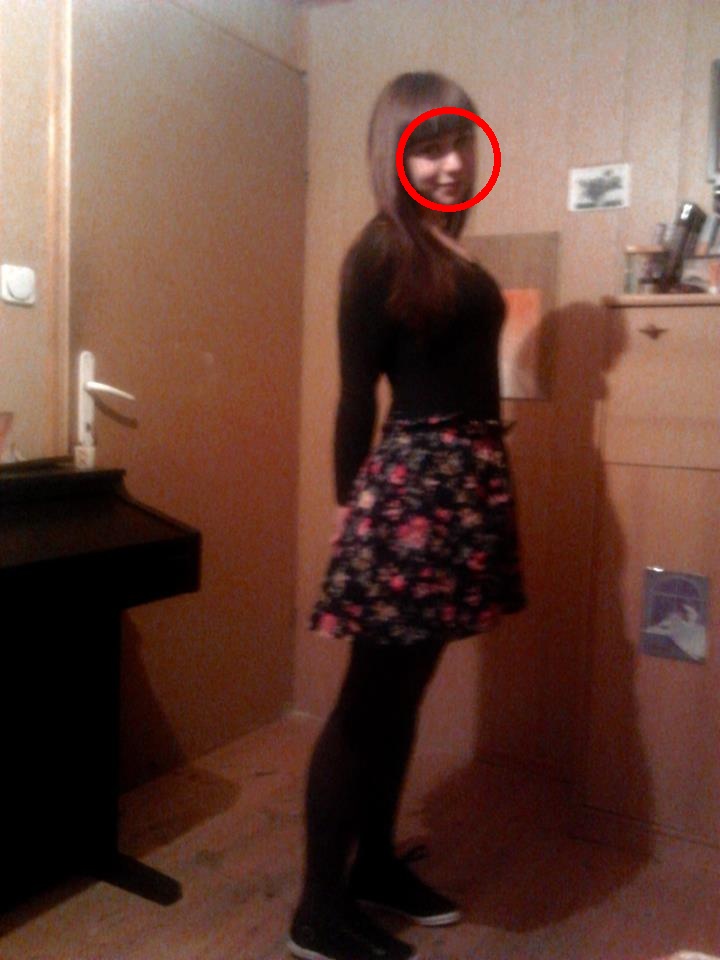}
				}

				\caption
				{
					Some results obtained by our system on real-world images.
				}
				\label{fig:detections}
			\end{figure*}

			Some qualitative results obtained by our system can be seen in Figure \ref{fig:detections}.
			A video demonstrating rotation invariant face detection is available at \url{http://youtu.be/1lXfm-PZz0Q}.
			Furthermore, for readers who wish to test the method themselves, demo applications can be downloaded from \url{http://public.tel.fer.hr/odet/}.
			Complete source code is provided at \url{https://github.com/nenadmarkus/pico}.

	\section{Conclusion}\label{sec:end}
		In this paper we have shown that an object detection system based on pixel intensity comparisons organized in decision trees can achieve competitive results at high processing speed.
		This is especially evident on devices with limited hardware support for floating point operations.
		This could prove important in the embedded system and mobile device industries because it can reduce hardware workload and prolong battery life.

		Further advantages of our method are:
		\begin{itemize}
			\item
				The method does not require the computation of integral images, image pyramid, HOG pyramid or any other similar data structure.
			\item
				All binary tests in internal nodes of the trees are based on the same feature type.
				For comparison, Viola and Jones used $5$ different types of Haar-like features to achieve their results.
			\item
				There is no need for image preprocessing prior to detection
				(such as contrast normalization, resizing, Gaussian smoothing or gamma correction).
			\item
				The method can easily be modified for fast detection of in-plane rotated objects.
		\end{itemize}

	\small
	{
		\section*{Acknowledgements}
			This research is partially supported by Visage Technologies AB, Link\"oping, Sweden, by the Ministry of Science, Education and Sports of the Republic of Croatia, grant number 036-0362027-2028 "Embodied Conversational Agents for Services in Networked and Mobile Environments" and by the European Union through ACROSS project, 285939 FP7-REGPOT-2011-1.
	}

	\small
	{
		\bibliographystyle{plain}
		\bibliography{references}
	}

\end{document}